\documentclass[conference,a4paper]{IEEEtran}
\IEEEoverridecommandlockouts
\usepackage{cite}
\usepackage{amsmath,amssymb,amsfonts}
\usepackage{algorithmic}
\usepackage{graphicx}
\usepackage{textcomp}
\usepackage{xcolor}
\usepackage{algorithmic}
\usepackage{url}
\usepackage{lipsum}
\usepackage{multirow}
\usepackage[normalem]{ulem}
\useunder{\uline}{\ul}{}
\usepackage{bbding}
\usepackage{pifont}
\usepackage{wasysym}
\usepackage[ruled,lined,boxed]{algorithm2e}
\DeclareMathAlphabet{\mathbcal}{OMS}{cmsy}{b}{n}
\def\BibTeX{{\rm B\kern-.05em{\sc i\kern-.025em b}\kern-.08em
    T\kern-.1667em\lower.7ex\hbox{E}\kern-.125emX}}
\begin{document}

\title{Transformer-based Joint Source Channel Coding for Textual Semantic Communication\\
\thanks{This work was supported in part by the Natural Science Foundation of China (NSFC) under Grant U2233216 and 62071044, in part by the Shandong Province Natural Science Foundation under Grant ZR2022YQ62, and in part by the Beijing Nova Program. Corresponding author: Zhen Gao.}
}
\author{\IEEEauthorblockN{Shicong~Liu\IEEEauthorrefmark{1},  Zhen~Gao\IEEEauthorrefmark{1}\IEEEauthorrefmark{2}\IEEEauthorrefmark{3}, Gaojie~Chen\IEEEauthorrefmark{4}, Yu~Su\IEEEauthorrefmark{5}, and Lu~Peng\IEEEauthorrefmark{5}
	}
	\IEEEauthorblockA{\IEEEauthorrefmark{1}MIIT Key Laboratory of Complex-Field Intelligent Sensing, Beiing Institute of Technology, Beijing 100081, China}
	\IEEEauthorblockA{\IEEEauthorrefmark{2}Advanced Research Institute of Multidisciplinary Science, Beijing Institute of Technology, Jinan 250307, China}
	\IEEEauthorblockA{\IEEEauthorrefmark{3}Yangtze Delta Region Academy of Beijing Institute of Technology, Beijing Institute of Technology, Jiaxing 314000, China}
	\IEEEauthorblockA{\IEEEauthorrefmark{4}Institute for Communication Systems, University of Surrey, Guildford, United Kingdom}
	\IEEEauthorblockA{\IEEEauthorrefmark{5}China Mobile (Chengdu) Institute of Research and Development, Chengdu, Sichuan, 610000, China}
	
	Email: \{scliu, gaozhen16\}@bit.edu.cn
}

\maketitle

\begin{abstract}
	The Space-Air-Ground-Sea integrated network calls for more robust and secure transmission techniques against jamming. In this paper, we propose a textual semantic transmission framework for robust transmission, which utilizes the advanced natural language processing techniques to model and encode sentences. Specifically, the textual sentences are firstly split into tokens using wordpiece algorithm, and are embedded to token vectors for semantic extraction by Transformer-based encoder. The encoded data are quantized to a fixed length binary sequence for transmission, where binary erasure, symmetric, and deletion channels are considered for transmission. The received binary sequences are further decoded by the transformer decoders into tokens used for sentence reconstruction. Our proposed approach leverages the power of neural networks and attention mechanism to provide reliable and efficient communication of textual data in challenging wireless environments, and simulation results on semantic similarity and bilingual evaluation understudy prove the superiority of the proposed model in semantic transmission.
\end{abstract}


\section{Introduction}

\IEEEPARstart{T}{he} rapid growth of mobile communication spurred significant research in capacity and coverage. Space-air-ground-sea integrated network (SAGSIN) has recently been envisioned as one of the promising network structures in future communication systems, which provides seamless coverage with unparalleled transmission rate\cite{sat}. However, the openness and predictable communication patterns such as fixed satellite ephemeris of SAGSIN are making it difficult to avoid jamming attacks. Therefore, robustness and security issues remains to be solved against jamming\cite{jamming1}.

Channel coding has been extensively employed in mobile communication systems to enhance the transmission performance against jamming, where the redundant bits inserted during the encoding procedure can effectively check and correct errors generated during transmission at the receiving end\cite{turbo,ldpc}. However, traditional communication systems tend to separate source and channel coding in their designs, with optimality reliant on infinite code length\cite{icassp}, which is particularly challenging to achieve in signaling exchange among heterogeneous massive nodes in SAGSIN\cite{sagsin}. Moreover, maintaining bit-level transmission accuracy in many goal-oriented scenarios, such as text, images, and video streams, becomes less necessary since minor errors at bit-level will not severely compromise the target of conveying the meaning.




Motivated by this, this paper proposes a joint source-channel coding (JSCC) scheme for textual data, which employs advanced natural language processing (NLP) models to encode sequentially arranged texts into fixed-length binary vectors. In the proposed approach, we no longer measure the bit-level accuracy of transmitted content; instead, we consider utilizing subjective semantic similarity evaluation models in~\cite{bleu} and~\cite{simcse} to assess the semantic transmission. Note that similar scenarios can be easily extended to signaling transmission among massive heterogeneous nodes in SAGSIN systems. In summary, the main contributions of this paper are as follows:
\begin{itemize}
	\item We propose a JSCC framework for textual semantic extraction and transmission based on advanced pre-trained NLP models\cite{attn,bert}. The encoder part is mainly built by pre-trained Bidirectional Encoder Representation from Transformers (BERT) model\cite{bert} with a fixed-length quantizer, which extracts the semantic meaning from texts and embeds the meanings into vectors.
	\item We use Transformer decoders\cite{attn} as recurrent textual sequence generator, which decodes the received bits and produce output sentences with minimized semantic loss.
	\item We adopt several binary channel such as binary erasure, symmetric, and deletion models to simulate the transmission scenario, and introduce a subjective semantic similarity evaluation model to evaluate the robustness and effectiveness of the proposed scheme.
\end{itemize}

\section{System Model}
Consider a JSCC scenario specified for textual data transmission, where the sentence ${\bf s}_I$ in English language is encoded directly with tokens\footnote{Tokenization is a fundamental concept in modern NLP, which breaks down the given sentences into word pieces (tokens).} as the minimum encoding unit. The sentences can therefore be expressed as
\begin{equation}
	{\bf s}_I = \left\{ w_1,w_2,\cdots,w_{L_I} \right\},
\end{equation}
where $w_i\in\mathcal{V}$ denotes the $i$th token in the vocabulary set $\mathcal{V}$, and $L_I$ denotes the number of tokens of the input sentence. The tokenized sentence is then mapped by a JSCC encoder to binary codewords as ${\bf C}=f_{\rm enc}\left( {\bf s}_I \right)\in\mathbb{B}^{L_E\times Q}$, where $L_E\geq L_I$ is the embedding length of ${\bf s}_I$, and $Q$ is the number of coding bits for each token. The binary codewords are passed into JSCC decoders after channel ${\bf h}\left(\cdot\right)$ as
\begin{equation}
	{\bf s}_O = f_{\rm dec}\left( {\bf h}\left( {\bf C} \right) \right)=\left\{ \hat{w}_1,\hat{w}_2,\cdots,\hat{w}_{L_O} \right\},
\end{equation}
where $\hat{w}_i\in{\mathcal{S}}$ is generated according the vocabulary set $\mathcal{S}$, and $L_O$ denotes the number of tokens in the output sentence ${\bf s}_O$.
As for channel ${\bf h}(\cdot)$, we consider several different kind of binary channel as
\begin{itemize}
	\item Binary erasure channel (BEC) ${\bf h}_{\rm BEC}(\cdot)$. The receiver either receives the transmitted bits correctly or with probability $P_e$ receives a message that the bit was not received.
	\item Binary symmetric channel (BSC) ${\bf h}_{\rm BSC}(\cdot)$. Similar to BEC, BSC either receives the transmitted bits correctly or with probability $P_e$ flips the received bits. 
	\item Deletion channel (DC) ${\bf h}_{\rm DC}(\cdot)$. The receiver either receives the transmitted bits correctly, or with probability $P_e$ loses the bits.
\end{itemize}
\section{Proposed Transformer-based Scheme}
In this section we introduce the proposed Transformer-based textual semantic coding scheme in details. 
\subsection{Encoder Part}
\begin{figure}[t]
	\centering
	\includegraphics[width=0.48\textwidth]{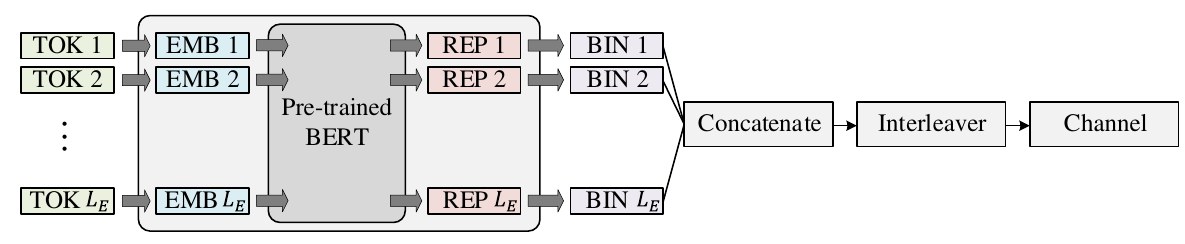}
	\caption{Encoder and decoder design for textual JSCC framework.}
	\label{fig:enc}
\end{figure}
To deal with work tokens in sentence ${\bf s}_I$, we firstly embed the textual contents into digital representations. In this paper, for the vocabulary set $\mathcal{V}$ with cardinality $N_V$, we define an embedding mapping $f_{\rm emb}: \mathcal{V}\rightarrow \mathbb{R}^{N_{\rm emb}}$, through which the tokens can be mapped into $N_{\rm emb}$-dimension vectors as
\begin{equation}
	{\bf W}_{\rm emb} = f_{\rm emb}\left(\left\{ w_1,\cdots,w_{L_I} \right\}\right)\in\mathbb{R}^{L_E\times N_{\rm emb}}.
\end{equation}
Rather than considering shallow window-based or global co-occurrence matrix-based\cite{icassp} methods such as word2vec and GloVe, we directly adopt the pre-trained Transformer-based word embedding model, which effectively solves the existing issues, e.g., out-of-vocabulary (OOV), in the aforementioned statistic embedding methods. Different from conventional encoding schemes, since we need to distinguish different segment of sentences, special tokens such as separator, mask, start and end of sentence indicators, etc, are added to the embedding matrix. Therefore, the embedding matrix of each sentence ${\bf W}_{\rm emb} = \left[ {\bf w}_1,\cdots, {\bf w}_{{L_E}} \right]^{\rm T}$ is usually longer\footnote{We encode the textual data in token level, therefore will not affect the transmission efficiency.} than the original textual sequence.

To fully extract the semantic information in the embedded sentences, the pre-trained BERT language model is adopted as semantic encoder. Unlike previous adopted BLSTM model\cite{icassp} that captures context in a sequential manner, Transformer-based models are able to use self-attention mechanism to capture context in parallel to cut down the computational overhead. For the $i$th encoder layer, input embedding matrix ${\bf W}_{\rm emb}^{(i)}$ is projected as
\begin{subequations}
	\begin{align}
		\mathbf{Q}^{i,h} & ={\bf W}_{\rm emb}^{(i)}{\bf W}_Q^{i,h}\in\mathbb{R}^{L_E\times N_{\rm attn}},\label{eq:q} \\
		\mathbf{K}^{i,h} & ={\bf W}_{\rm emb}^{(i)}{\bf W}_K^{i,h}\in\mathbb{R}^{L_E\times N_{\rm attn}},\label{eq:k} \\
		\mathbf{V}^{i,h} & ={\bf W}_{\rm emb}^{(i)}{\bf W}_V^{i,h}\in\mathbb{R}^{L_E\times N_{\rm attn}},\label{eq:v}
	\end{align}
	\label{selfattn1}%
\end{subequations}
where ${\bf W}_Q^{i,h}$, ${\bf W}_K^{i,h}$, and ${\bf W}_V^{i,h}$ are respectively the projection matrix for query, key, and value at the $i$th layer, $h=1,\cdots, N_{\rm h}$ denotes the index of $N_{\rm h}$ parallel attention chunks, and $N_{\rm emb} = N_{\rm attn}N_{\rm h}$. For each attention chunk we have 
\begin{equation}
	{\bf A}_{i,h} = {\rm Softmax}\left(\frac{{\bf Q}^{i,h}\left({\bf K}^{i,h} \right)^{\rm T}}{\sqrt{N_{\rm attn}}}\right){\bf V}^{i,h}\in\mathbb{R}^{L_E\times N_{\rm attn}},
	\label{selfattn2}
\end{equation}
which can further contribute to multi-head attention by
\begin{equation}
	{\bf O}_i = \left[{\bf A}_{i,1},\cdots,{\bf A}_{i,N_{\rm h}}\right]{\bf W}_{O}^{(i)}\in\mathbb{R}^{L_E\times N_{\rm emb}}.
	\label{selfattn3}
\end{equation}
Calculations in Eqs.~\eqref{selfattn1}-\eqref{selfattn3} form the multi-head attention. Linear projection operations with residual connection are inserted between two self-attention operations as 
\begin{equation}
	{\bf W}_{\rm emb}^{(i+1)} = \sigma\left( \sigma\left( {\bf O}_i{\bf W}_{i,1}+{\bf B}_{i,1} \right){\bf W}_{i,2}+{\bf B}_{i,2} \right)+{\bf O}_i,
\end{equation}
where ${\bf W}_{i,k}$ and ${\bf B}_{i,k}$ ($k=1,2$) are respectively the weight and bias matrix for linear projection, and $\sigma(\cdot)$ is the non-linear activation function.

After $M_{\rm enc}$ layers of encoders, the extracted semantic information matrix is dimensionally reduced as ${\bf C} = {\bf W}_{\rm emb}^{(M_{\rm enc})} {\bf W}_{\rm out}\in\mathbb{R}^{L_E\times Q}$, and is further binarized for transmission. The binarizer maps the entries in ${\bf C}$ by ${\rm tanh}(\cdot)$, and quantizes each entry with hard threshold $0$ to $\{-1,1\}$. For end-to-end training procedure, the gradient can be passed backward via straight-through method\cite{VQVAE}.

\subsection{Decoder Part}
\begin{figure}[t]
	\centering
	\includegraphics[width=0.40\textwidth]{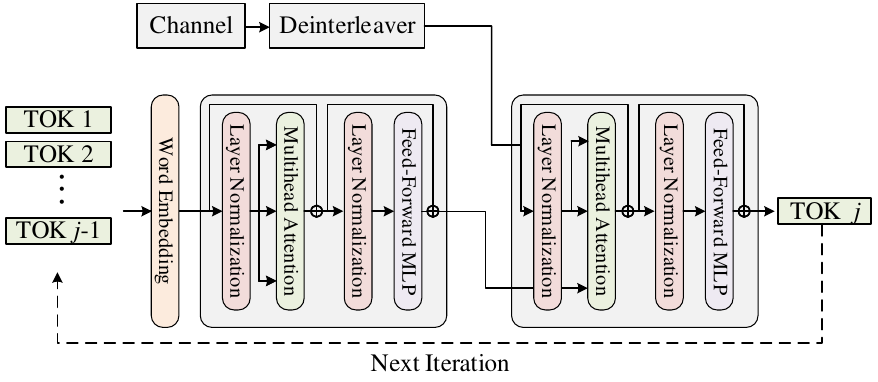}
	\caption{Decoder design for textual JSCC framework.}
	\label{fig:dec}
\end{figure}
Given the received bits through error-prone channel\footnote{For deletion channel ${\bf h}_{\rm DC}(\cdot)$, the received bits are firstly zero-padded to $\mathbb{B}^{L_E\times Q}$ before sent to the decoder.} as $\hat{\bf C} = {\bf h}({\bf C})=\left[ \hat{\bf c}_1, \cdots, \hat{\bf c}_{L_E} \right]^{\rm T}\in\mathbb{B}^{L_E\times Q}$, the generative decoder infers the output probability of each token recursively after linear projection ${\bf W}_{\rm dec} = \hat{\bf C}{\bf W}_{\rm up}=[{\bf w}_1,\cdots,{\bf w}_{L_E}]^{\rm T}\in\mathbb{R}^{L_E\times N_{\rm emb}}$. For the $i$th decoder layer ($i=1,\cdots,M_{\rm dec}$), the $j$th output token is generated in two steps:
\begin{itemize}
	\item Self-attention. The previous generated tokens $\{\hat{\bf w}^{(i-1)}_k \}_{k=1}^{j-1}$ are concatenate as 
	\begin{equation}
		\hat{\bf W}_{\rm in}^{(i)} = \left[\hat{\bf w}^{(i)}_1,\cdots,\hat{\bf w}^{(i)}_{j-1}\right]^{\rm T}\in\mathbb{R}^{j-1\times N_{\rm emb}},
		\label{eq:gen}
	\end{equation}
	which is then used for self-attention calculation similar to Eqs.~\eqref{selfattn1}-\eqref{selfattn3}.
	\item Cross-attention. Instead of using the generated tokens, the received tokens ${\bf W}_{\rm dec}$ are used for a cross-attention with generated tokens. Specifically, ${\bf W}_{\rm dec}$ is used to calculate query and key matrix by Eq.~\eqref{eq:q} and \eqref{eq:k}, while $\hat{\bf W}_{\rm in}^{(i)}$ is used for value calculation in Eq.~\eqref{eq:v}. Note that the first dimension of $\hat{\bf W}_{\rm in}^{(i)}$ is padded to the same length as ${\bf W}_{\rm dec}$ before Eqs.~\eqref{selfattn2} and \eqref{selfattn3}.
\end{itemize}

The aforementioned steps together form the $i$th decoding layer. 
During the training phase, the teacher-forcing technique is adopted, which means that the generated tokens in Eq.~\eqref{eq:gen} are forced to be the ground-truth tokens to avoid error accumulation.

\subsection{Channel Model}
As we have mentioned before, three kinds of channel models are considered to evaluate the transmission. To implement the channel model in the end-to-end training model, neural network components are fully utilized to achieve different channel models. For BEC model with erasure probability $P_e$ that fails to receive each bit, the dropout layer\cite{icassp} can be adopted as
\begin{equation}
	\hat{\bf C} = {\bf h}_{\rm BEC}\left( {\bf C};P_e \right) = {\rm dropout}\left( {\bf C} ;P_e \right),
\end{equation}
where $P_e$ denotes the probability for each entry in ${\bf C}$ to be erased as $0$, hence the received symbol set is $\{-1,0,1\}$.

For BSC model with flipping probability $P_e$, the received bits can be described as
\begin{equation}
	\begin{aligned}
		\hat{\bf C} =& {\bf h}_{\rm BSC}\left( {\bf C} \right)\\ 
		=& 2\:{\rm dropout}\left( {\bf C} ;P_e \right) -{\bf C}.
	\end{aligned}
\end{equation}

For DC model with deletion probability $P_e$, the shape of received bits $\hat{\bf C}$ may change from time to time. In our implementations, $\hat{\bf C}$ is padded to the same shape with received bit through other channel models.
\subsection{Datasets and Metrics}



The ability to convey the complexity and nuance of language is paramount for successful semantic communication, and auto-regressive datasets may struggle to make language models capture this essential aspect. They still validate the model through bit-level correctness by supervising the output using the input data. Therefore, we consider to build a large rephrasing dataset for \textit{short} conversation scenarios. Rephrasing dataset is built by aggregating the following open datasets:
\begin{itemize}
	\item Multi30K\cite{multi30k}: A collection of around $30,000$ image-caption pairs in multiple languages, where the pictures depict daily life scenes. Most of the image captions are short sentences, which is suitable for our goal.
	\item Microsoft COCO\cite{mscoco}: An even larger collection of around $330,000$ images and their associated $5$ captions each.
	\item WikiAnswers\cite{wikianswers}: The author voted similar questions dataset that groups up the similar questions.
\end{itemize}

Since Multi30K dataset has no similar meaning text pairs in one language, we augment it using T5 model\cite{T5} by rephrasing the English image captions. The aggregated dataset hence has two main components, the auto-regressive part that supervises the output texts by the same input texts, and the semantically similar part that supervise the output texts by the rephrased texts. 
Besides, multiple evaluation criteria are introduced to verify the performance of semantic encoding as
\begin{itemize}
	\item Perplexity (PPL), which is calculated by taking the inverse probability of the test set, measures how confused the model is when it encounters unseen data. It is $e$ to the power of cross entropy of the prediction probability distribution, and literally represents the the average number of candidate words in each prediction step. 
	\item Bilingual evaluation understudy (BLEU)\cite{bleu} works by comparing the $n$ contiguous sequences of words in the machine-generated text with those in the reference text, and assigning a score based on the overlap between them. 
	\item Similarity. BERT-based sentence embedding uses a pooling token followed by a linear projection, which shows unstable performance in evaluation the semantic similarity. In this paper, we introduce SimCSE model\cite{simcse} to generate sentence embeddings using contrastive learning techniques to evaluate the similarity between them. 
\end{itemize}

\section{Main Results}
In this section, we compare the proposed Transformer-based semantic encoding approach with conventional source and channel coding baselines.

\subsection{Simulation Setup}
\begin{figure}[t]
	\centering
	\includegraphics[width=0.4\textwidth]{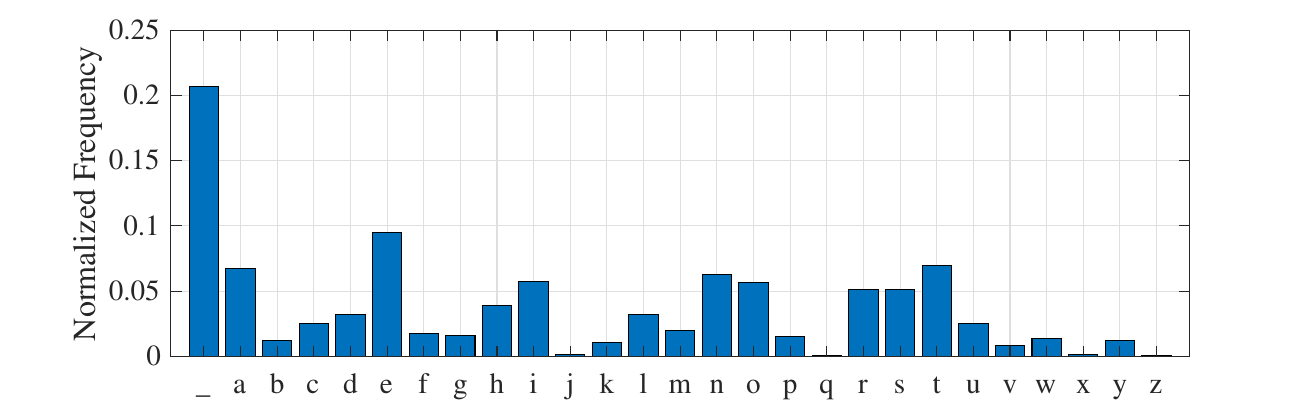}
	\caption{Normalized occurrence frequency of letters and space (i.e., `\_') in short conversation dataset.}
	\label{fig:pdf}
\end{figure}
The encoder part is composed of a pre-trained BERT model and a binarizer, where we embed the tokens into vectors with dimension $N_{\rm emb}=768$ and sent to $M_{\rm enc} = 6$ encoder layers. The number of parallel attention modules is $N_{\rm h} = 12$, and the quantization bits is set according to the baseline methods discussed below. The decoder part has the same parameters with encoder part, and the channel parameter $P_e$ varies during the simulation. The overall dataset has around $300,000$ short sentence pairs, and is iterated for $20$ epochs for training.

Throughout the simulation, we consider a weighted-sum loss function with PPL, BLEU, and semantic similarity, where PPL shows the dominate weight among all metrics.

\subsection{Numerical Results}

\begin{figure}[t]
	\centering
	\includegraphics[width=0.4\textwidth]{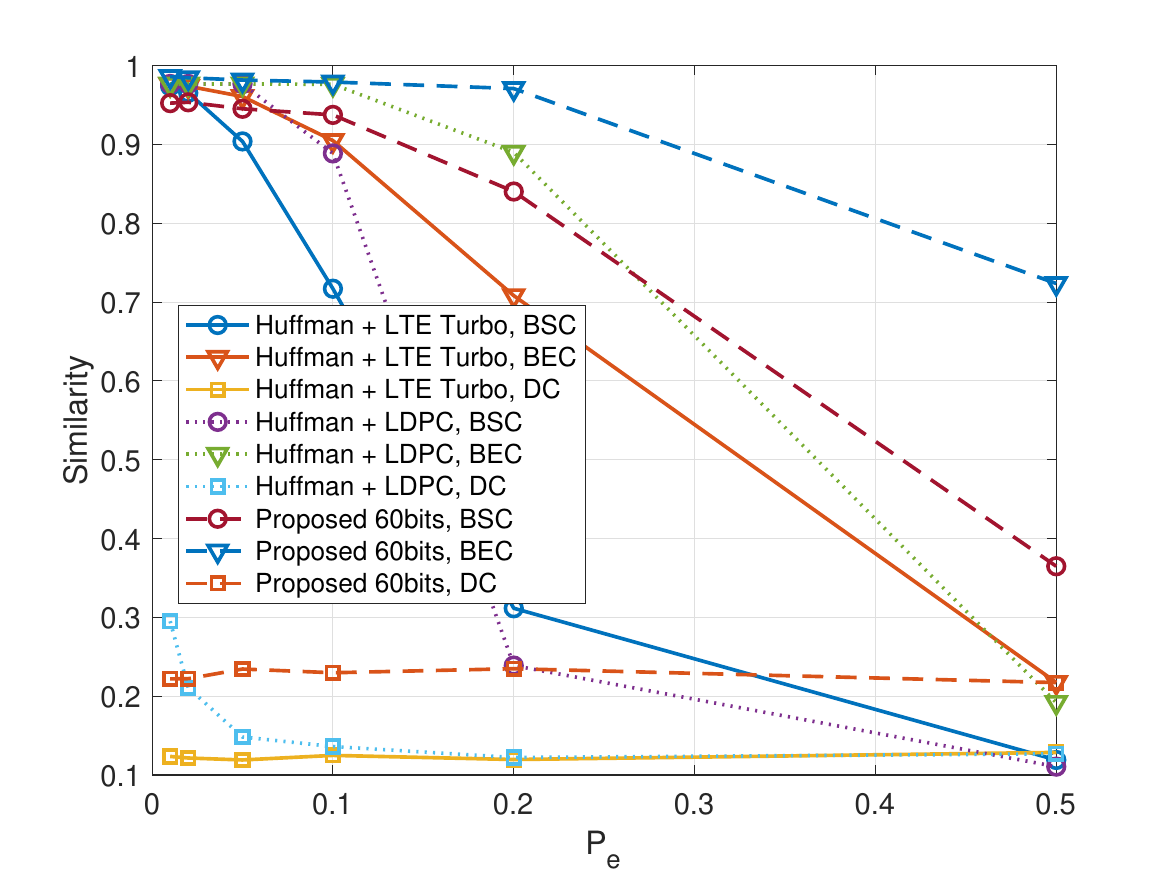}
	\caption{Semantic similarity performance for our proposed scheme and LTE standards under different binary channels.}
	\label{fig:sim}
\end{figure}

The coding overhead is firstly considered. For conventional coding schemes as baselines, we consider using Huffman coding as source coding method, and Turbo/LDPC coding as channel coding method. Given the symbol distribution statistics (without considering punctuation marks in short contexts) in Fig.~\ref{fig:pdf}, the source entropy can be calculated as $H=-\sum_{i=1}^{27}p_i\log_2(p_i) = 4.059$ bits per symbol, which is approximated by Huffman coding with average length $\bar{H} = 4.093$ bits per symbol. The channel coding baselines are Turbo and LDPC code standardized by LTE, with $R = 1/3$ code rate configuration. Therefore, the average coding length for each symbol is around $\bar{H}/R = 12.279$ bits, and considering the average length for each token is around $\bar{L} = 4.588$ letters, the considered quantization bits for each token is set around $Q = \lceil \bar{L}\bar{H}/R \rceil = 60$ bits.
\begin{table*}[!t]
	\centering
	\caption{The source and received text of the proposed scheme under different channel parameters $P_e$.}
	\label{tab1}
	\begin{tabular}{c|c|c|l|c}
		\hline
		Case           & Error Rate              & Channel & Texts     & Flag                                                                       \\ \hline
		\multirow{10}{*}{A} & $P_e=0$                    &         & two bikes ride through a state park filled with campers and other bikers  &-       \\ \cline{2-5} 
		& \multirow{3}{*}{$P_e=0.1$} & BEC     &  two bikes ride through a state park filled with campers and other bikers  &\Checkmark       \\ \cline{3-5} 
		&                         & BSC     &  two bikes ride through a state park filled with campers and other bikers   &\Checkmark      \\ \cline{3-5} 
		&                         & DC      &   a man in a blue shirt and a black hat is standing in front of a building.  &\XSolidBrush      \\ \cline{2-5} 
		& \multirow{3}{*}{$P_e=0.2$} & BEC     &  two bikes ride through a state park filled with campers and other bikers  &\Checkmark        \\ \cline{3-5} 
		&                         & BSC     &  two bike riders are riding a \textbf{beautiful} park filled with campers and other bikers &{\Checkmark}\textsuperscript{{\kern-0.82em\ding{55}}} \hspace{-2pt}\\ \cline{3-5} 
		&                         & DC      &  a man is standing in front of a large building.        &\XSolidBrush                          \\ \cline{2-5} 
		& \multirow{3}{*}{$P_e=0.5$} & BEC     &  two \textbf{motorcycle} in a state park filled with camp \textbf{camp boats} and other bikers  &{\Checkmark}\textsuperscript{{\kern-0.82em\ding{55}}}     \\ \cline{3-5} 
		&                         & BSC     &  the knit is descending a golf yard with a camp with aqua top and reds.  &\XSolidBrush         \\ \cline{3-5} 
		&                         & DC      &  a man in a red shirt and a black hat is standing in front of a building.  &\XSolidBrush       \\ \hline
		\multirow{10}{*}{B} & $P_e=0$                    &         &  an old man playing the violin is being watched by a child.  &    -                 \\ \cline{2-5} 
		& \multirow{3}{*}{$P_e=0.1$} & BEC     &  an old man playing the violin is being watched by a child  &\Checkmark                       \\ \cline{3-5} 
		&                         & BSC     &  an old man playing the violin is being watched by a child  &\Checkmark                       \\ \cline{3-5} 
		&                         & DC      &  a man is standing on a skateboard in the air.    &\XSolidBrush                                \\ \cline{2-5} 
		& \multirow{3}{*}{$P_e=0.2$} & BEC     &  an old man playing the violin is being watched by a child    & \Checkmark                    \\ \cline{3-5} 
		&                         & BSC     &  an old man playing the violin is being watched by a child  & \Checkmark                       \\ \cline{3-5} 
		&                         & DC      &  a man is standing in front of a large building.  & \XSolidBrush                                \\ \cline{2-5} 
		& \multirow{3}{*}{$P_e=0.5$} & BEC     &  an old man playing the violin is \textbf{chased} by a child   & {\Checkmark}\textsuperscript{{\kern-0.82em\ding{55}}}                             \\ \cline{3-5} 
		&                         & BSC     &  a woman playing a cigarette in a boatful of three toddler   & \XSolidBrush                     \\ \cline{3-5} 
		&                         & DC      &  a group of people are standing in front of a building.  & \XSolidBrush                         \\ \hline
		\multirow{10}{*}{C} & $P_e=0$                    &         &  an elephant whose ear flapped across a road.  &    -                 \\ \cline{2-5} 
		& \multirow{3}{*}{$P_e=0.1$} & BEC     &  an elephant with its ear flapped up crossing a road.  &\Checkmark                       \\ \cline{3-5} 
		&                         & BSC     &  an elephant whose ear flapped across a road.  &\Checkmark                       \\ \cline{3-5} 
		&                         & DC      &  a group of people are standing on a beach.    &\XSolidBrush                                \\ \cline{2-5} 
		& \multirow{3}{*}{$P_e=0.2$} & BEC     &  an elephant with its ear flapped up crossing a road.    & \Checkmark                    \\ \cline{3-5} 
		&                         & BSC     &  an elephant with its ear flapped up is crossing a road.  & {\Checkmark}                       \\ \cline{3-5} 
		&                         & DC      &  a group of people are standing in front of a building.  & \XSolidBrush                                \\ \cline{2-5} 
		& \multirow{3}{*}{$P_e=0.5$} & BEC     &  an elephant is \textbf{standing on a road flying overhead}.   & {\Checkmark}\textsuperscript{{\kern-0.82em\ding{55}}}                             \\ \cline{3-5} 
		&                         & BSC     &  there is a portrait displaying a leave poster on a mountain.   & \XSolidBrush                     \\ \cline{3-5} 
		&                         & DC      &  a group of people are standing in front of a building.  & \XSolidBrush                         \\ \hline
	\end{tabular}
\end{table*}

The similarity evaluated by SimCSE model under different binary channel is given in Fig.~\ref{fig:sim}. As we can see, all of the coding schemes show better performance under BEC channel model, which is followed by BSC channel model, while for the DC channel model all of the schemes fail to achieve the communication goal. Among these schemes, the proposed Transformer-based joint coding scheme achieves the best performance under all three channel models, even when the binary channel flips or erases $P_e=20\%$ of the transmitted bits. 

\begin{figure}[t]
	\centering
	\includegraphics[width=0.4\textwidth]{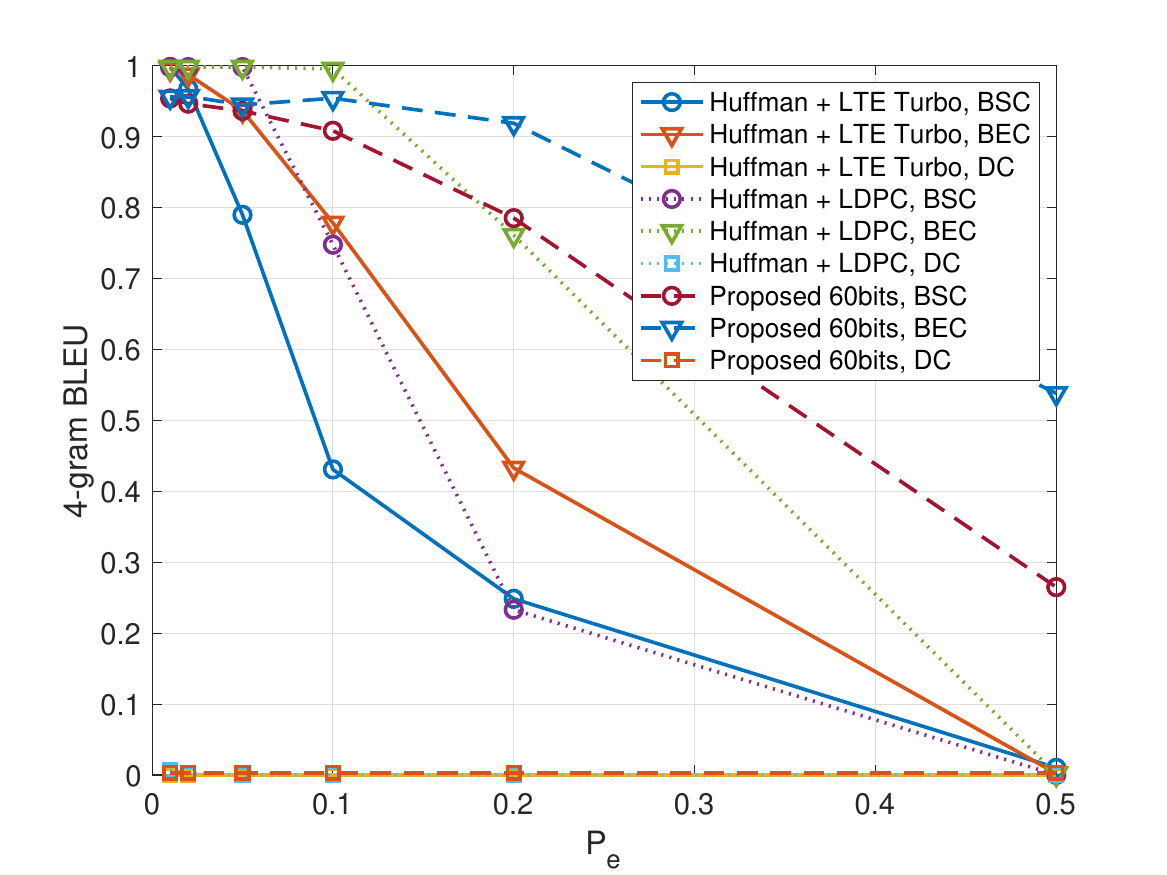}
	\caption{4-gram BLEU performance for our proposed scheme and LTE standards under different binary channels.}
	\label{fig:bleu}
\end{figure}

We further study the BLEU performance of the proposed method under the aforementioned channel. As is shown in Fig.~\ref{fig:bleu}, since no bit error is detected for LDPC coding schemes when $P_e$ is low, the BLEU performance of LDPC schemes are higher than the proposed methods under low flipping/erasure rate $P_e$. However, when channel states get worse, as shown in Fig.~\ref{fig:bleu} when $P_e = 20\%$, the proposed method maintains a high BLEU performance, which indicates the potential of the proposed method with lower coding length $Q$. As the rate $P_e$ continue increasing to $50\%$, all of the conventional methods fail to transmit any of the words from the source, while the proposed method still reconstruct several words at the receiver, which reveals the superiority of proposed joint coding method.

Some of the typical outputs of our proposed model is also given in the Table~\ref{tab1}. The results are produced under very high error rate starting from $P_e=0.1$, which is critical even for Turbo and LDPC coding schemes with coding rate $1/3$ as shown before. However, in short conversation scenarios the coding length $Q=60$ bits of proposed scheme is able to provide large enough coding space for each sentences against the lossy channel. For BEC and BSC scenarios with $P_e = 0.1$ and $0.2$, the source texts can be accurately reconstructed at the receiver, while for DC scenarios the model fails to reconstruct the original texts. Note that for case A and B, the model tends to provide the exactly same sentences at the receiver, while for case C the model is more likely to give sentences with similar meanings. 
Since the decoder acts as a generative model, which produce results given the query sequence $\hat{\bf C}$, the decoded sentences after deletion channel can fallback to several fixed sentences as shown in the table.

It is also worth mentioning that the conventional methods are producing incorrect spellings and bad punctuation marks after severe lossy channel, which makes it difficult to read at the receiver, and is not shown in the table here.


\section{Conclusion}
This paper proposed a Transformer-based JSCC scheme for textual semantic transmission tasks, which showed performance superiority against conventional LTE Turbo and LDPC channel coding schemes with Huffman source coding. The semantic similarity and BLEU performance also indicated that the proposed joint coding scheme has the potential to transmit semantic information under lower coding length $Q$. However, to produce the variable coding length against different channel states $P_e$ and achieve semantic transmission under deletion channel are still challenges to be solved in the future.

\end{document}